\documentclass[10pt,twocolumn,letterpaper]{article}

\usepackage{iccv}
\usepackage{times}
\usepackage{epsfig}
\usepackage{graphicx}
\usepackage{amsmath}
\usepackage{amssymb}
\usepackage{enumitem}
\usepackage{booktabs}
\usepackage{slashbox}
\graphicspath{ {./} }

\usepackage[breaklinks=true,bookmarks=false]{hyperref}
\usepackage[all]{hypcap}

\iccvfinalcopy 

\def\iccvPaperID{2029} 

\ificcvfinal\fi

\begin{document}

\title{Improving Pedestrian Attribute Recognition With \\ Weakly-Supervised Multi-Scale Attribute-Specific Localization}

\author{Chufeng Tang$^{1}$ \ \ Lu Sheng$^{2}$ \ \ Zhaoxiang Zhang$^{3}$ \ \ Xiaolin Hu$^{1}$\thanks{Corresponding author.} \\
$^{1}$State Key Laboratory of Intelligent Technology and Systems, \\
Institute for Artificial Intelligence, Department of Computer Science and Technology, \\
Beijing National Research Center for Information Science and Technology, Tsinghua University \\
$^{2}$College of Software, Beihang University \
$^{3}$Institute of Automation, Chinese Academy of Sciences \\
{\tt\small \{tcf18@mails,xlhu@mail\}.tsinghua.edu.cn \ lsheng@buaa.edu.cn \ zhaoxiang.zhang@ia.ac.cn}
}

\maketitle
\ificcvfinal\thispagestyle{empty}\fi

\begin{abstract}
  Pedestrian attribute recognition has been an emerging research topic in the area of video surveillance.
  To predict the existence of a particular attribute, it is demanded to localize the regions related to the attribute. However, in this task, the region annotations are not available. How to carve out these attribute-related regions remains challenging.
  Existing methods applied attribute-agnostic visual attention or heuristic body-part localization mechanisms to enhance the local feature representations, while neglecting to employ attributes to define local feature areas.
  We propose a flexible Attribute Localization Module (ALM) to adaptively discover the most discriminative regions and learns the regional features for each attribute at multiple levels.
  Moreover, a feature pyramid architecture is also introduced to enhance the attribute-specific localization at low-levels with high-level semantic guidance.
  The proposed framework does not require additional region annotations and can be trained end-to-end with multi-level deep supervision.
  Extensive experiments show that the proposed method achieves state-of-the-art results on three pedestrian attribute datasets, including PETA, RAP, and PA-100K.
\end{abstract}

\section{Introduction} \label{sec:intro}

Recognition of pedestrian attributes, \eg gender, age, and clothing style, has drawn extensive attention because of its great potential in video surveillance applications, such as face verification \cite{facever}, person retrieval \cite{retrieval2,retrieval1}, and person re-identification \cite{layne2012pedestrian,Peng2016JointL,wang2018transferable}.
Recently, methods based on the Convolutional Neural Networks (CNN) \cite{resnet,bn} achieve great success in pedestrian attribute recognition by learning powerful features from images.
Some existing works \cite{deepmar,sudowe2015person} treat pedestrian attribute recognition as a multi-label classification problem and extract feature representations only from the whole input images.
%
These holistic methods usually rely on global features, but regional features are more significant for fine-grained attribute classification.

\begin{figure}[t]
\begin{center}
  \includegraphics[width=0.85\linewidth]{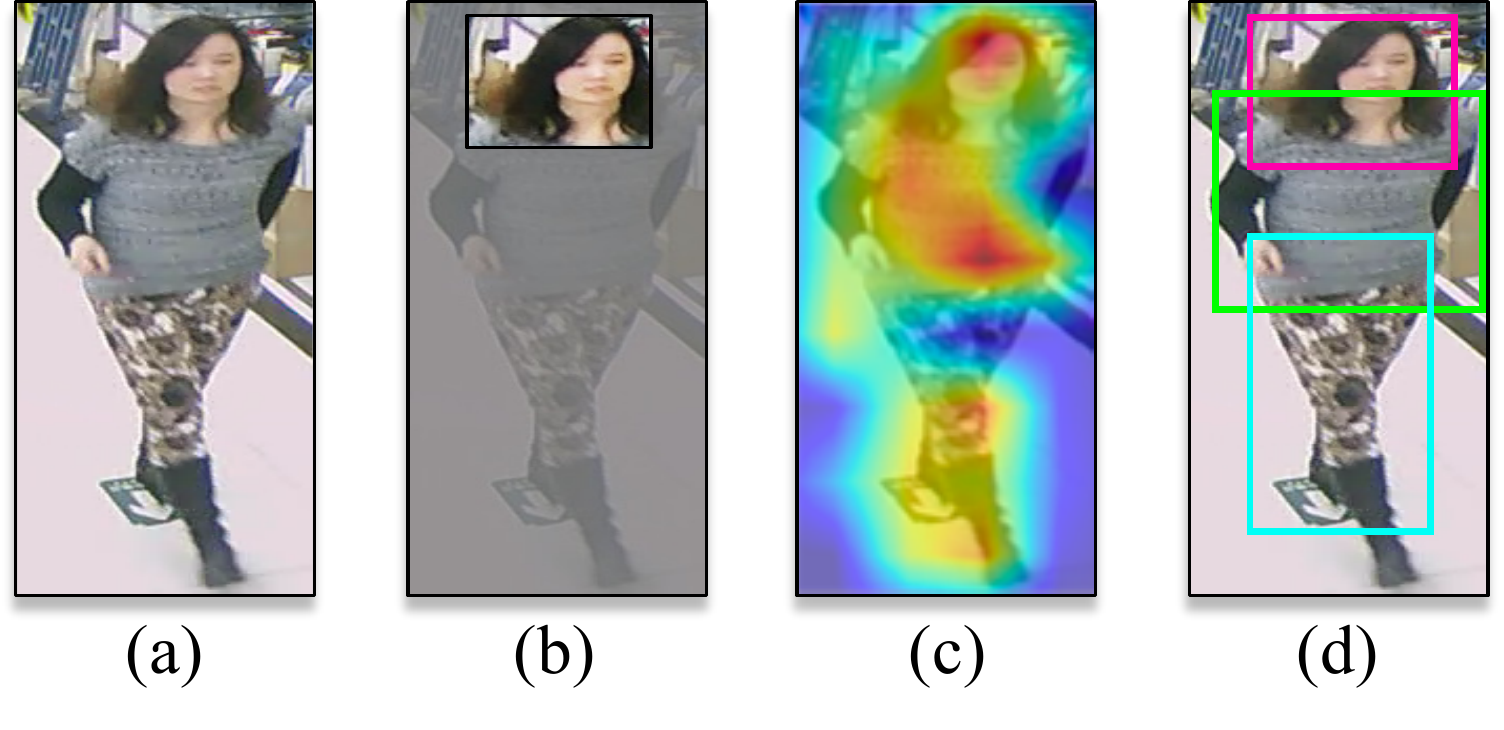}
\end{center}
\vspace{-4mm}
   \caption{Attentive regions generated by different methods when recognizing the attribute \textit{Longhair}.
   (a) The original input image.
   (b) Attribute-specific region generated by our proposed method, which is indeed localized into a head-related region.
   (c) Attention mask generated by attribute-agnostic attention methods \cite{hpnet,deepimb,zhuspatialreg}, which covers a broad region but not specific to \textit{Longhair}.
   (d) Body parts generated by part-based methods \cite{li2018pose,liu2018localization,yang2016attribute,zhang2014panda}, which extract features from these body parts.}
\label{fig:pedestrian}
\vspace{-2mm}
\end{figure}

Intuitively, attributes can be localized into some relevant regions in a pedestrian image.
As illustrated in Figure~\ref{fig:pedestrian} (b), when recognizing~\emph{Longhair}, it is reasonable to focus on the head-related regions.
Recent methods attempt to leverage the attention localization to promote learning discriminative features for attribute recognition.
A popular solution \cite{hpnet,deepimb,zhuspatialreg} is to employ the visual attention mechanism to capture the most relevant features.
These methods usually generate attention masks from certain layers and then multiply them to corresponded feature maps so as to extract the attentive features.
However, it is ambiguous which mask encodes a given attribute's location, and there is no specific mechanism that guarantees the correspondences between attributes and attention masks.
As shown in Figure \ref{fig:pedestrian} (c), the learned attention mask attends to a broad region which is not specific to the required attribute~\emph{Longhair}.
An alternative way is to leverage predefined rigid parts \cite{zhu2015multi} or external part localization modules \cite{li2018pose,liu2018localization,yang2016attribute,zhang2014panda}.
Some works apply body-parts detection \cite{zhang2014panda}, pose estimation \cite{li2018pose,yang2016attribute} and region proposals \cite{liu2018localization} to learn part-based local features. 
As shown in Figure \ref{fig:pedestrian} (d), these methods extract local features from the localized body parts (\eg head, torso, and legs).
However, most of them just fuse the part-based features with global features, which still fail to indicate the attribute-region correspondence but require extra computational resources for sophisticated part localization.
%

Different from these methods, we propose a flexible \textit{Attribute Localization Module} (ALM) that can automatically discover the discriminative regions and extract region-based feature representations in an attribute-specific manner.
Specifically, the ALM consists of a tiny channel-attention sub-network to fully exploit the inter-channel dependencies of the input features, followed by a spatial transformer \cite{stn} to localize the attribute-specific regions adaptively.
%
Moreover, we embed multiple ALMs at different feature levels and introduce a feature pyramid architecture by integrating high-level semantics to reinforce the attribute localization at low-levels.
In addition, ALMs at different feature levels are trained by the same set of attribute supervisions, called \textit{deep supervision} \cite{lee2015deeply,wang2018resource}, where the final predictions are obtained through a voting scheme to output the maximum responses across different feature levels.
This voting scheme will suggest a best prediction occurs in one feature level that has the most accurate attribute region, without interference of negative features from inappropriate regions.
The proposed framework is end-to-end trainable and requires only image-level annotations.
The contributions of this work can be summarized as follows:
\setlist{nolistsep}
\begin{itemize}[noitemsep]
  \item We propose an end-to-end trainable framework which performs attribute-specific localization at multiple scales to discover the most discriminative attribute regions in a weakly-supervised manner.
  \item We propose a feature pyramid architecture by leveraging both low-level details and high-level semantics to enhance the multi-scale attribute localization and region-based feature learning in a mutually reinforcing manner. The multi-scale attribute predictions are further fused by an effective voting scheme.
  \item We conduct extensive experiments on three publicly available pedestrian attribute datasets (PETA \cite{deng2014pedestrian}, RAP \cite{rap}, and PA-100K \cite{hpnet}) and achieve significant improvement over the previous state-of-the-art methods.
\end{itemize}

\begin{figure*}[t]
  \begin{center}
    \includegraphics[width=0.85\linewidth]{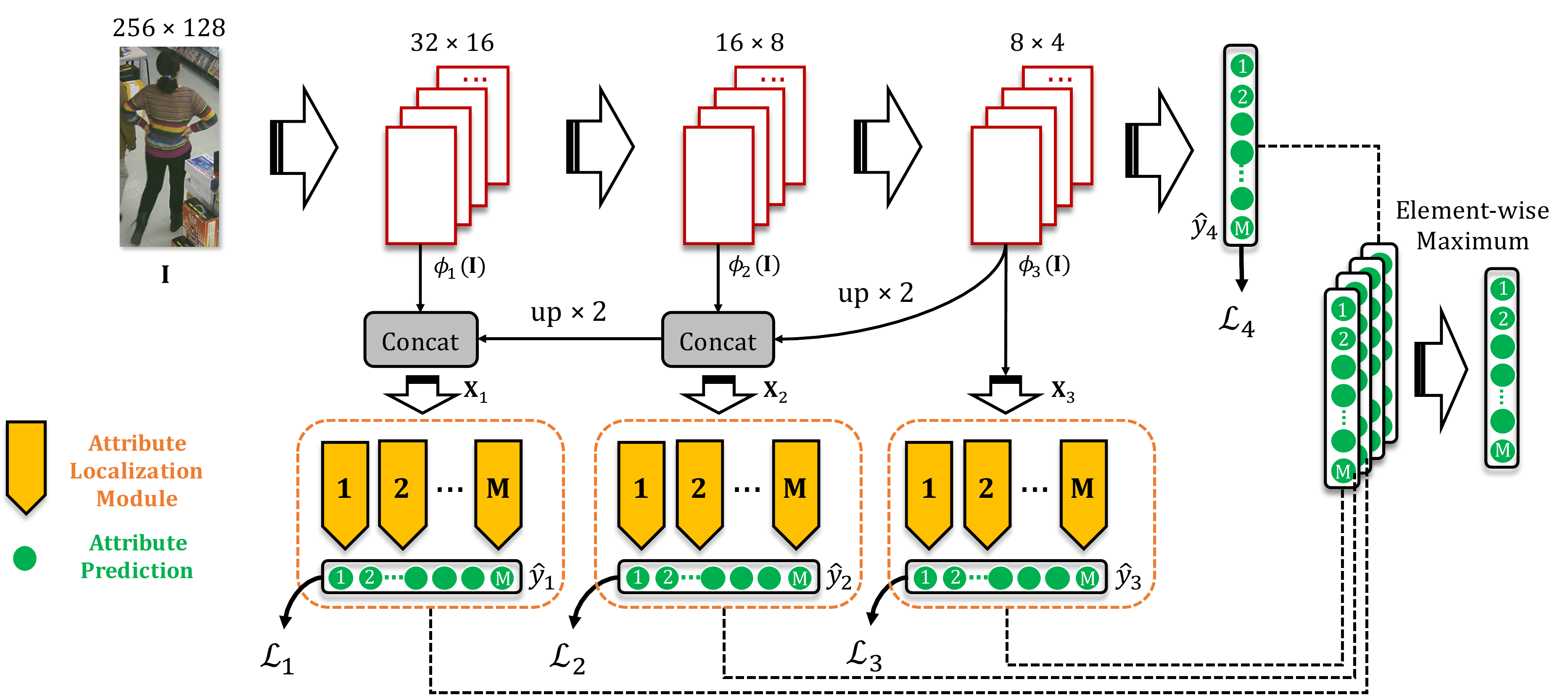}
  \end{center}
  \vspace{-1mm}
     \caption{Overview of the proposed framework.
     The input pedestrian image is fed into the main network with both bottom-up and top-down pathways.
     Features combined from different levels are fed into multiple \textit{Attribute Localization Modules} (Figure \ref{fig:ALM}),
     which perform attribute-specific localization and region-based feature learning.
     Outputs from different branches are trained with \textit{deep supervision} and aggregated through an element-wise maximum operation for inference.
     $M$ is the total number of attributes. Best viewed in color.}
  \label{fig:framework}
\end{figure*}

\begin{figure}[t]
  \begin{center}
    \includegraphics[width=1.02\linewidth]{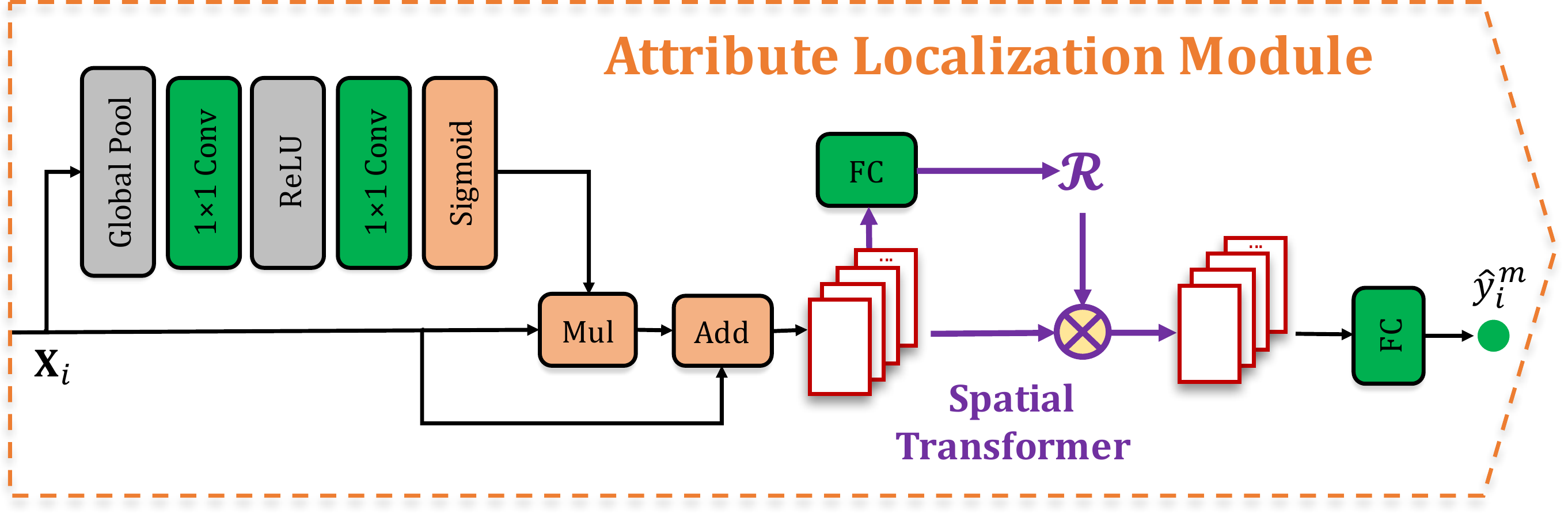}
  \end{center}
  \vspace{-2mm}
     \caption{Details of the proposed \textit{Attribute Localization Module} (ALM), which consists of a tiny channel-attention sub-network and a simplified spatial transformer.
     The ALM takes the combined features $\mathbf{X}_{i}$ as input and produces an attribute-specific prediction.
     Each ALM only serves one attribute at a singe level.}
  \label{fig:ALM}
\end{figure}

\section{Related Works} \label{sec:related-works}
\textbf{Pedestrian Attribute Recognition}.
Earlier pedestrian attribute recognition methods \cite{deng2014pedestrian,layne2012pedestrian,zhu2013pedestrian} rely on hand-crafted features such as color and texture histograms, and trained separately.
However, the performance of these traditional methods is far from satisfactory.
More recently, methods based on the Convolutional Neural Networks achieved great success in pedestrian attribute recognition.
Wang \etal \cite{wang2019PARSurvey} give a brief review of these methods.
Sudowe \etal \cite{sudowe2015person} propose a holistic CNN model to jointly learn different attributes.
Li \etal \cite{deepmar} formulate pedestrian attribute recognition as a multi-label classification problem and propose an improved cross-entropy loss function.
However, the performance of these holistic methods is limited due to the lack of consideration of the prior information in attributes.
Some recent approaches attempt to exploit the spatial relations and semantic relations among attributes to further improve the recognition performance.
These methods can be classified into three basic categories:
(1) \textbf{Relation-based}:
Some works \cite{wang2017attribute,zhao2018grouping} exploit semantic relations to assist attribute recognition.
Wang \etal \cite{wang2017attribute} propose a CNN-RNN based framework to exploit the interdependency and correlation among attributes.
Zhao \etal \cite{zhao2018grouping} divide the attributes into several groups and attempt to explore the intra-group and inter-group relationships.
However, these methods require manually defined rules, \eg prediction order, attribute group, which are hard to determine in real applications.
(2) \textbf{Attention-based}:
Some researchers \cite{hpnet,deepimb,deepview,zhuspatialreg} introduce the visual attention mechanism in attribute recognition.
Liu \etal \cite{hpnet} propose a multi-directional attention model to learn multi-scale attentive features for pedestrian analysis.
Sarafianos \etal \cite{deepimb} extend the spatial regularization module \cite{zhuspatialreg} to learn effective attention maps at multiple scales.
Although recognition accuracy has been improved, these methods are attribute-agnostic and fail to take the attribute-specific information into consideration.
(3) \textbf{Part-based}: The part-based methods usually extract features from some localized body-parts.
Zhu \etal \cite{zhu2015multi} divide the whole image into 15 rigid patches and fuse features from different patches.
Yang \etal \cite{yang2016attribute} and Li \etal \cite{li2018pose} leverage external pose estimation module to localize body-parts.
Liu \etal \cite{liu2018localization} also explore attribute regions in a weakly supervised manner while they assign attribute regions to some fixed proposals generated by EdgeBoxes \cite{Zitnick2014EdgeBL} in advance, which is not fully-adaptive and end-to-end trainable.
These methods rely either on predefined rigid parts or on sophisticated part localization mechanisms, which are less robust to pose variances and require extra computational resources.
By contrast, the proposed method localizes the most discriminative regions in an attribute-specific manner, which is not considered in most of the existing works.

\textbf{Weakly Supervised Attention Localization}.
In addition to pedestrian attribute recognition, the idea of performing attention localization without region annotations is also extensively investigated in other visual tasks.
Jaderberg \etal \cite{stn} propose the well-known \textit{Spatial Transformer Network} (STN) which can extract attentional regions with any spatial transformation in an end-to-end trainable manner.
Some recent works \cite{li2017learning,li2018harmonious} adopt STN to localize body-parts for person re-identification.
Fu \etal \cite{fu2017look} attempt to recursively learn discriminative region for fine-grained image recognition.
Wang \etal \cite{wang2017multi} search the discriminative regions with STN and LSTM for multi-label classification, while not in a label-specific manner.
The proposed method is inspired by these works but can adaptively localize the individual informative regions for each attribute.

\textbf{Feature Pyramid Architecture}.
There are several works exploiting top-down or skip connections that incorporate features across levels, \eg U-Net \cite{ronneberger2015u}, Stacked hourglass network \cite{newell2016stacked}.
The proposed feature pyramid architecture is similar to \textit{Feature Pyramid Networks} (FPN) \cite{fpn}, which have been studied in various object detection and segmentation models \cite{shrivastava2016beyond,zhu2018bidirectional}.
To the best of our knowledge, this work is the first attempt of employing these ideas to localize attentive regions for pedestrian attribute recognition.


\section{Proposed Method}

The overview of the proposed framework is illustrated in Figure \ref{fig:framework}.
As shown, the proposed framework consists of a main network with feature pyramid structures, and a group of \textit{Attribute Localization Modules} (ALM) applied to different feature levels.
The input pedestrian image is first fed into the main network without additional region annotations, and a prediction vector is obtained at the end of the bottom-up pathway.
The details of ALM are shown in Figure \ref{fig:ALM}.
Each ALM only perform attribute localization and region-based feature learning for one attribute at a single feature level.
The ALMs at different feature levels are trained in a \textit{deep supervision} manner.
Formally, given an input pedestrian image $\mathbf{I}$ along with its corresponding attribute labels $y = \left[ y ^ { 1 } , y ^ { 2 } , \ldots , y ^ { M } \right] ^ { T }$ where $M$ is ths total number of attributes in the dataset and $y ^ { m },m \in 1 , \dots , M$ is a binary label that indicates the presence of the $m$-th attribute if $y ^ { m }=1$, and $y ^ { m }=0$ otherwise.
We adopt the BN-Inception \cite{bn} architecture as the backbone network in our framework. In principle, the backbone can be replaced with any other CNN architecture.
Implementation details are shown in Appendix \ref{appendix_a}.

\subsection{Network Architecture}
The key idea of this work is to perform attribute-specific localization for improving attribute recognition.
It is well known that features in deeper CNN layers have coarser resolutions.
Even though we can precisely localize the attribute regions based on semantically stronger features, it is still difficult to extract region-based discriminative features since some finer details may disappear.
In contrast, features in lower layers always capture rich details but poor contextual information, resulting in unreliable attribute localization.
Obviously, low-level details and high-level semantics are complementary to each other.
Therefore, we propose a feature pyramid architecture, inspired by the FPN alike models \cite{fpn,zhu2018bidirectional}, to enhance the attribute localization and region-based feature learning in a mutually reinforcing manner.
As illustrated in Figure \ref{fig:framework}, the proposed feature pyramid architecture consists of a bottom-up pathway and a top-down pathway.

The bottom-up pathway, implemented by BN-Inception network, consists of multiple \verb'inception' blocks with different feature levels.
In this paper, we conduct attribute localization with bottom-up features generated from three different levels: the \verb'incep_3b', \verb'incep_4d', and \verb'incep_5b' block respectively, where they have strides of $\{ 8, 16, 32 \}$ pixels with respect to the input image.
The selected \verb'inception' blocks are both at the end of their corresponded stages, where blocks of the same stage keep the same feature maps resolution, since we believe the last block should have strongest features.
Given an input image $\mathbf{I}$, we denote the bottom-up features generated from the above blocks as $\phi_{i}(\mathbf{I}) \in \mathbb{ R }^{H_{i} \times W_{i} \times C_{i}}, i \in \{ 1,2,3 \} $.
For $256 \times 128$ RGB input images, the spatial size $H_{i} \times W_{i}$ equal to $32 \times 16$, $16 \times 8$, and $8 \times 4$ respectively.

In addition, the top-down pathway contains three lateral connections and two top-down connections, as shown in Figure \ref{fig:framework}.
The lateral connections are simply used to reduce the dimensionalities of bottom-up features to $d$, where $d=256$ in our implementation.
The higher level features are transmitted through the top-down connections and meanwhile go through an upsampling operation.
Afterward, features from adjacent levels are concatenated as follows:
\begin{equation}
  \mathbf{X}_{i} = \{ f(\phi_{i}(\mathbf{I})) , g(\mathbf{X}_{i+1}) \}, i \in \{ 1,2 \},
\end{equation}
where $f$ is a $1 \times 1$ convolutional layer for dimensionality reduction, $g$ refers to upsampling with nearest neighbor interpolation.
Since the highest level features have no top-down connection, we only conduct dimensionality reduction for $\phi_{3}(\mathbf{I})$:
\begin{equation}
  \mathbf{X}_{3} = f(\phi_{3}(\mathbf{I})).
\end{equation}
The channel size of $\mathbf{X}_{i}$ equal to $d,2d,3d$ for $i \in \{ 1,2,3 \}$.
The combined features $\mathbf{X}_{i}$ are used for attribute-specific localization.

\subsection{Attribute Localization Module} \label{sec:ALM}
As mentioned in Section \ref{sec:intro}, several existing methods attempt to extract local features through attribute-agnostic visual attention, predefined rigid parts or external part localization modules.
However, these methods are not the optimal solution since they overlook the significance of attribute-specific localization.
As shown in Figure \ref{fig:pedestrian} (c,d), attentive regions belong to different attributes are mixed together, which is inconsistent with the original intention that narrowing the attentive region for improving attribute recognition.
We believe that attribute-specific localization is a better choice since it can disentangle the confused attention masks into several individual regions, where each region for a specific attribute.
Moreover, the learned attribute-specific regions are more interpretable since we can observe the attribute-region correspondence intuitively.
What we need is a mechanism that can learn an individual bounding box, representing the discriminative region, in feature maps for a given attribute.
The well-known RoI pooling technique \cite{girshick2015fast} is inappropriate since it requires region annotations, which are not available in pedestrian attribute datasets.
Inspired by the recent success of \textit{Spatial Transformer Network} (STN) \cite{stn}, we propose a flexible \textit{Attribute Localization Module} (ALM) to automatically discover the discriminative regions for each attribute in a weakly-supervised manner.
The overview of the proposed ALM is illustrated in Figure \ref{fig:ALM}.

As shown, each ALM contains a spatial transformer layer originates from STN.
STN is a differentiable module which is capable of applying a spatial transformation to a feature map, \eg cropping, translation, and scaling.
In this paper, we adopt a simplified version of STN since we treat the attribute region as a simple bounding box, which can be realized through the following transformation:
\begin{equation}
  \left( \begin{array} { c } { x _ { i } ^ { s } } \\ { y _ { i } ^ { s } } \end{array} \right) = \left[ \begin{array} { c c c } { s _ { x } } & { 0 } & { t _ { x } } \\ { 0 } & { s _ { y } } & { t _ { y } } \end{array} \right] \left( \begin{array} { c } { x _ { i } ^ { t } } \\ { y _ { i } ^ { t } } \\ { 1 } \end{array} \right),
\end{equation}
where $s_{x}$, $s_{y}$ are scaling parameters, and $t_{x}$, $t_{y}$ are translation parameters, the expected bounding box can be obtained through these four parameters.
$\left( x _ { i } ^ { s } , y _ { i } ^ { s } \right)$ and $\left( x _ { i } ^ { t } , y _ { i } ^ { t } \right)$ are the source coordinates and target coordinates of the $i$-th pixel.
To some extent, this simplified spatial transformer can be viewed as a differentiable RoI pooling, which is end-to-end trainable without region annotations.
To accelerate the convergence, we simply constrain $s_{x}$,$s_{y}$ to $(0,1)$ and $t_{x},t_{y}$ to $(-1,1)$ by a \textit{sigmoid} and \textit{tanh} activation, respectively.

In addition, we also introduce a tiny channel-attention sub-network, as shown in Figure \ref{fig:ALM}.
As mentioned above, the ALM takes the features combined from adjacent levels as input, where both finer details and strong semantics take the same proportion (both have $d$ channels), which means they equally contribute to attribute localization.
However, the expected proportion should vary from attribute to attribute.
For example, more details should be paid when recognizing finer attributes.
Therefore, we introduce this channel-attention sub-network, similar to SE-Net \cite{hu2018squeeze}, to modulate the inter-channel dependencies.

Specifically, the input features $\mathbf{X}_{i}$ pass through a series of linear and nonlinear layers, producing a weight vector for feature recalibration across channels.
The reweighted features are obtained by channel-wise multiplying the weight vector with $\mathbf{X}_{i}$, and an extra residual link is applied to preserve the complementary information.
Subsequently, a fully-connected layer is applied to estimate the transformation matrix, denoted as $\mathcal{R}$, and then the region-based features sampled by bilinear interpolation are used for attribute classification.
We simply formulate the prediction belong to $m$-th attribute at $i$-th level as:
\begin{equation}
  \hat y^{m}_{i} = ALM_{i}^{m}(\mathbf{X}_{i}).
\end{equation}

\subsection{Deep Supervision} \label{sec:deepsupervision}
As illustrated in Figure \ref{fig:framework}, four individual prediction vectors are obtained from three ALM groups and one global branch.
We apply the \textit{deep supervision} \cite{lee2015deeply,wang2018resource} mechanism for training where the four individual predictions are directly supervised by ground-truth labels.
During inference, multiple prediction vectors are aggregated through an effective voting scheme that producing the maximum responses across different feature levels.
The intuition behind this design is that each ALM should directly take the feedback about whether the localized region is accurate.
If we only preserve the supervision of the fused predictions (maximum or averaging), the gradients are not informative enough of how each level performs, such that some branches are trained insufficiently.
The maximum voting scheme is applied to choose the best predictions from different levels with the most accurate attribute region.

Specifically, we adopt the weighted binary cross-entropy loss function \cite{deepmar} at each stage, formulated as follow:
\begin{equation}
  \begin{split}
    \mathcal{L}_{i}(\hat y_{i}, y) = -\frac{1}{M} \sum_{m=1}^{M} \gamma^{m}(\,y^{m} \log (\sigma (\hat y^{m}_{i})) \\ + (1 - y^{m}) \log (1 - \sigma (\hat y^{m}_{i}))\,),
  \end{split}
\end{equation}
where $\gamma^{m} = e ^ { - a _ { m } }$ is the loss weight for $m$-th attribute and $a _ { m }$ is the prior class distribution of $m$-th attribute,
$M$ is the number of attributes,
$i$ represents the $i$-th branch, where $i \in \{1,2,3,4\}$,
and $\sigma$ refers to the \textit{sigmoid} activation.
The total training loss is calculated by summing over the four individual loss: $\mathcal{L} = \sum_{i=1}^{4} \mathcal{L}_{i}$.

\section{Experiments}

\subsection{Datasets and Evaluation Metrics}

The proposed method is evaluated on three publicly available pedestrian attribute datasets:
(1) The \textbf{PETA} dataset \cite{deng2014pedestrian} consists of 19,000 images with 61 binary attributes and 4 multi-class attributes.
Following the previous works \cite{deng2014pedestrian,deepview}, the whole dataset is randomly partitioned into three subsets: 9,500 for training, 1,900 for verification and 7,600 for testing.
We choose 35 attributes which the positive ratio is higher than $5\%$ for evaluation.
(2) The \textbf{RAP} dataset \cite{rap} contains 41,585 images which are
collected from 26 indoor surveillance cameras, where each image is annotated with 72 fine-grained attributes.
Following the official protocol \cite{rap}, we
split the whole dataset into 33,268 training images and 8,317 test images.
Only 51 binary attributes with the positive ratio higher than $1\%$ are selected for evaluation.
(3) The \textbf{PA-100K} dataset \cite{hpnet} is to-date the largest dataset for pedestrian attribute
recognition, which contains 100,000 pedestrian images in total collected from outdoor surveillance
cameras. Each image is annotated with 26 commonly used attributes. According to the official
setting \cite{hpnet}, the whole dataset is randomly split into 80,000 training images, 10,000 validation
images and 10,000 test images.

We adopt two types of metrics for evaluation \cite{rap}:
(1) Label-based: we calculate the mean accuracy (\textbf{mA}) as the mean of positive accuracy and negative
accuracy for each attribute. The mA criterion can be formulated as:
\begin{equation}
  mA = \frac { 1 } { 2N } \sum _ { i=1 } ^ { M } \left( \frac { TP _ { i } } { P _ { i } } + \frac { TN _ { i } } { N _ { i } } \right),
\end{equation}
where $N$ is the number of examples and $M$ is the number of attributes; $P _ { i }$ and $TP _ { i }$ are the number of positive examples and correctly predicted positive examples of the $i$-th attribute respectively; $N _ { i }$ and $TN _ { i }$ are defined similarly.
(2) Instance-based: we adopt four well-known criteria: \textbf{accuracy}, \textbf{precision}, \textbf{recall} and \textbf{F1 score}, details are omitted.

\subsection{Effectiveness of Critical Components}

%
As shown in Table \ref{Tab:component}, starting with the BN-Inception baseline, we gradually append each component and meanwhile compare it with several variants.
(1) \textbf{Attribute Localization Module}: We first evaluate the contribution of the simplified ALM (without channel-attention sub-network) by embedding ALMs at the final layer (\verb'incep_5b').
The increased mA and F1 scores demonstrate the effectiveness of attribute-specific localization.
Based on this fact, we further embed multiple ALMs at different feature levels (\verb'incep_3b,4d,5b'), and a greater improvement is achieved ($3.1\%$ and $1.3\%$ in mA and F1, respectively).
Considering the model complexity, we limit the number of levels to three in our framework.
(2) \textbf{Top-down Guidance}: Secondly, we evaluate the impact of the proposed feature pyramid architecture by comparing with three variants, which are different in how to combine features from different levels.
The first one is implemented by element-wise adding the features from different levels, like the original FPN \cite{fpn}, but the performance decreases.
The poor results suggest that some essential information may disappear if we disregard the feature mismatching problem.
The improved concatenation version achieves better results (improves $1.0\%$ in mA), which shows the success of high-level top-down guidance.
Moreover, the introduced channel-attention sub-network further improves mA a lot to $80.61\%$ by modulating the inter-channel dependencies.
(3) \textbf{Deep Supervision}: As mentioned in Section \ref{sec:deepsupervision}, the obtained gradients with only the supervision of fused predictions are not informative enough of how each level performs, while some branches are trained insufficiently.
To address this problem, ALMs at different levels are trained with \textit{deep supervision} mechanism.
For inference, the experimental results suggest that element-wise maximum is a superior ensemble method than averaging since some weaker existences are ignored in averaging.

\begin{figure}[t]
  \begin{center}
    \includegraphics[width=1.0\linewidth]{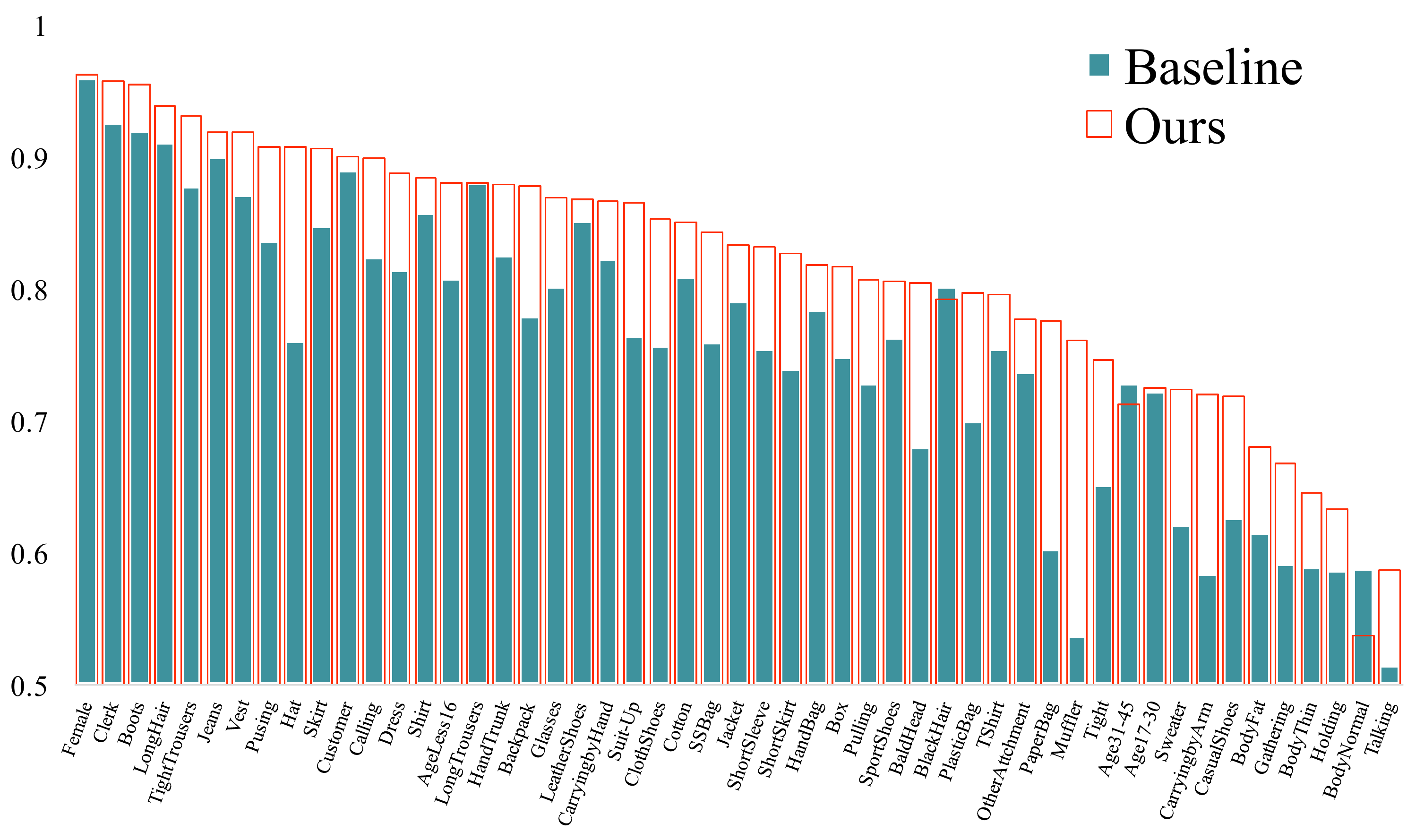}
  \end{center}
  \vspace{-3mm}
     \caption{Attribute-wise mA comparison on RAP dataset between our proposed method and the baseline model.
     The bars are sorted in descending order according to the larger mA between the two models.
     We can observe significant improvements on some fine-grained attributes, \eg \textit{BaldHead}, \textit{Hat} and \textit{Muffler}.}
  \label{fig:each_attribute}
\end{figure}

Removing all ALMs while keeping others unchanged results in a significant drop (last row in Table \ref{Tab:component}), which further confirmed the effectiveness of ALMs.
Compared with the baseline, the final model achieves a remarkable performance, improving $6.1\%$ and $1.9\%$ in mA and F1 metrics, respectively.
Figure \ref{fig:each_attribute} shows the attribute-wise mA comparison between the proposed method and baseline model on RAP dataset.
As shown, the proposed method achieves significant improvement on a number of attributes, especially some fine-grained attributes, \eg \textit{BaldHead}($23.1\%$), \textit{Hat}($12.4\%$) and \textit{Muffler}($13.5\%$).
The accurate recognition of these attributes shows the effectiveness of the proposed attribute-specific localization module.

\begin{table}[t]
  \small
  \begin{center}
  \begin{tabular}{l|cc}
    \hline
    \backslashbox{Component}{Metric} &mA &F1 \\ \hline
    Baseline &75.76 &78.20 \\ \hline\hline
    ALM at Single Level (5b) &77.45 &79.14 \\
    \textbf{ALM} at Multiple Levels (3b,4d,5b) &78.89 &79.50 \\ \hline\hline
    Top-down (Addition) &78.51 &79.42 \\
    Top-down (Concatenation) &79.93 &79.91 \\
    \textbf{Top-down} (Channel Attention) &80.61 &79.98 \\ \hline\hline
    Deep Supervision (Averaging) &80.70 &80.04 \\
    \textbf{Deep Supervision} (Maximum) (\textbf{Ours}) &\multicolumn{1}{l}{\textbf{81.87}} &\multicolumn{1}{l}{\textbf{80.16}}\\ \hline\hline
    \textbf{Ours} w/o ALMs & 78.91 & 79.55 \\\hline
  \end{tabular}
  \end{center}
  \caption{Performance comparisons on RAP dataset when gradually adding each proposed component to the baseline model (except the last row).
  Variants of the same component lie in the same group.
  \textbf{Bold} means the setting adopted in our final framework.}
  \label{Tab:component}
\end{table}

\begin{figure}[t]
  \begin{center}
    \includegraphics[width=1.0\linewidth]{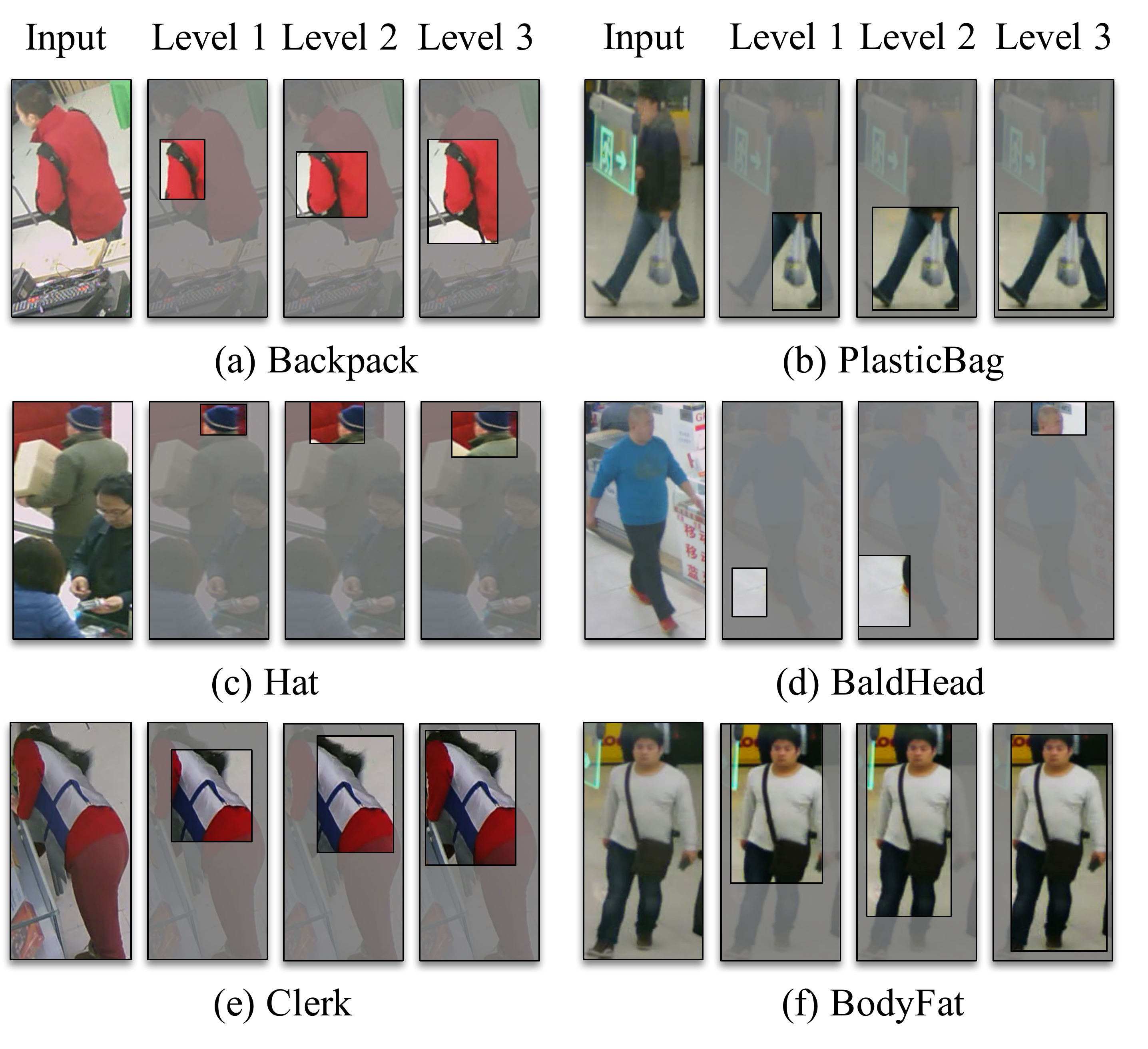}
  \end{center}
  \vspace{-4mm}
     \caption{Visualization of attribute localization results at different feature levels. Best viewed in color.}
  \label{fig:region}
  \vspace{-2mm}
\end{figure}

\subsection{Visualization of Attribute Localization}
Through the above quantitative evaluation, we can observe significant improvements on some fine-grained attributes.
In this subsection, we visualize the localized attribute regions from different feature levels for qualitative analysis.
In our implementation, the attribute regions are located within the feature maps, while the correspondence between a feature map pixel and an image pixel is not unique.
For a relatively coarse visualization, we simply map a feature-level pixel to the center of the receptive field on the input image, like SPPNet \cite{he2015spatial}.
As shown in Figure \ref{fig:region}, we display several examples belong to six different attributes, covering both abstract and concrete attributes.
As we can see, the proposed ALMs can successfully localize these concrete attributes, \eg \textit{Backpack}, \textit{PlasticBag}, and \textit{Hat}, into the corresponded informative regions, despite the extreme occlusions (a, c) or pose variances (e).
While recognizing the more abstract attributes \textit{Clerk} and \textit{BodyFat}, the ALMs tend to explore the larger regions, since they often require high-level semantics from the whole image.
In addition, a failure case is also provided, as shown in Figure \ref{fig:region}(d).
The ALMs fail to localize the expected regions at two lower levels when recognizing \textit{BaldHead}.
We believe that this problem originates from the highly imbalanced data distribution, where only $0.4$ percent of images are annotated with \textit{BaldHead} in the RAP dataset.
Although these localized attribute regions are relatively coarse, it is still acceptable for recognizing attributes because they indeed capture these most discriminative regions with large overlap.

\subsection{Different Attribute-Specific Methods} \label{sec4_4}

The most significant contribution of this work is the idea of localizing an individual informative region for each attribute, which we called \textbf{attribute-specific} and was not well investigated in previous works.
In this subsection, we conduct experiments to demonstrate the advantages of our proposed method by comparing with other attribute-specific localization methods, such as visual attention and predefined parts.
Different from the attribute-agnostic attention masks and body-parts illustrated in Figure \ref{fig:pedestrian}, we extend them to an attribute-specific version for comparison.
Firstly, we replace the proposed ALM with a spatial attention module while keeping others unchanged for a fair comparison.
In detail, we generate individual attention masks for each attribute through a global cross-channel averaging layer and a $3 \times 3$ convolutional layer, like HA-CNN \cite{li2018harmonious}.
For another comparison model, we divide the whole image into three rigid parts (head, torso, and legs) and extract part-based features with an RoI pooling layer, then manually define the attribute-part relations, \eg recognizing \textit{hat} only from the head part.
More details about the compared methods are shown in Appendix \ref{appendix_b}.
Experimental results are listed in Table \ref{Tab:mask}.
As expected, the proposed method largely outperforms the other two methods (improving $5.3\%$ and $3.5\%$ in mA, respectively).

\begin{figure}[t]
  \begin{center}
    \includegraphics[width=1.0\linewidth]{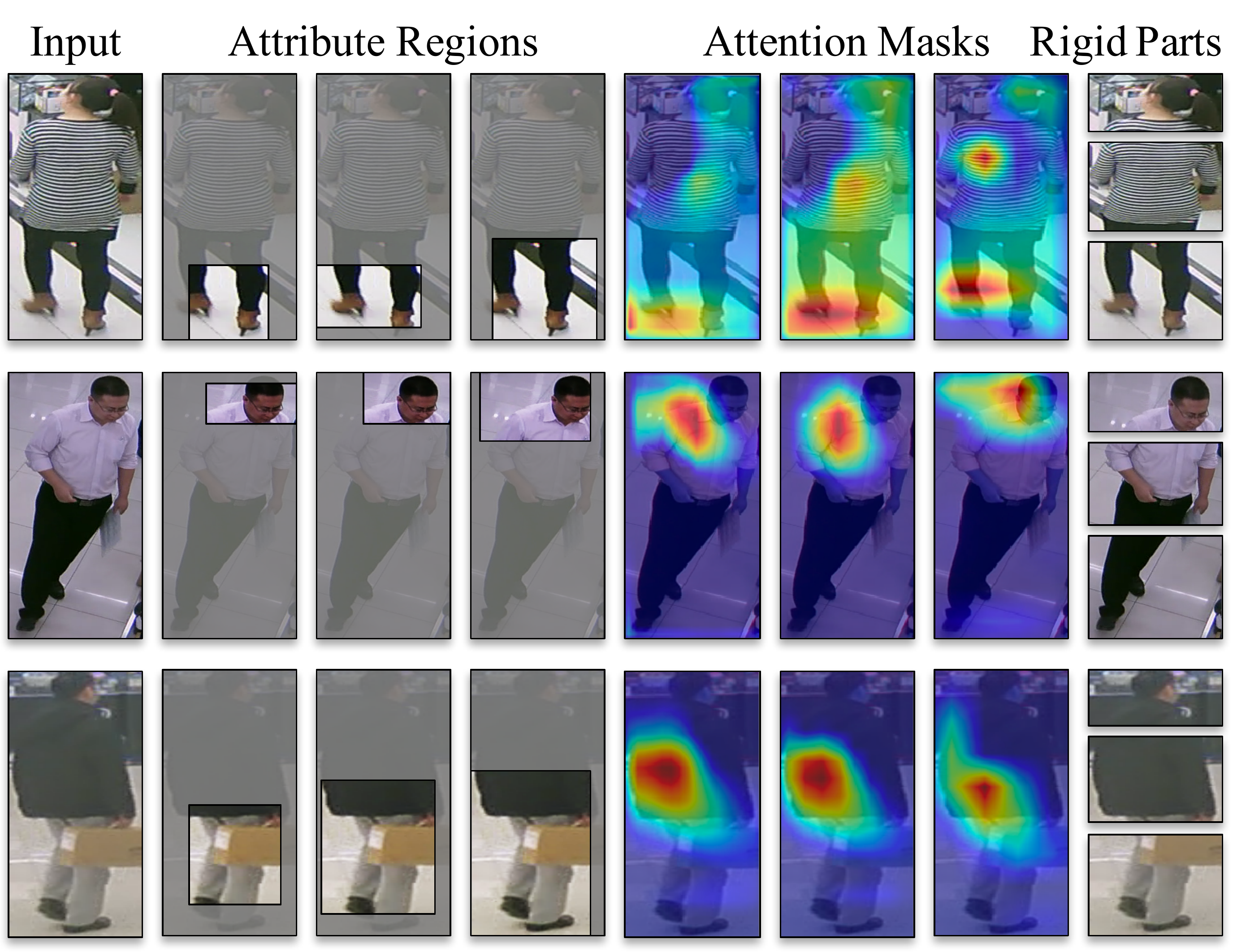}
  \end{center}
  \vspace{-3mm}
     \caption{Case studies of different attribute-specific localization methods on three different attributes: \textbf{\textit{Boots}} (Top), \textbf{\textit{Glasses}} (Middle), and \textbf{\textit{Box}} (Bottom).
      Different from Figure \ref{fig:pedestrian}, the attention masks and body-parts are applied in an attribute-specific manner.}
  \vspace{-3mm}
  \label{fig:mask}
\end{figure}

To better understanding the differences, we visualize these localization results in Figure \ref{fig:mask}.
As we can see, the attribute regions generated by ALMs are the most accurate and discriminative one.
Although the attention-based model achieves a not-bad result, the generated attention masks may attend to the irrelevant or biased regions.
While recognizing \textit{Box}, the attention masks fail to cover the expected regions, and we also observed that they tend to localize almost the same regions wherever the boxes are.
By contrast, the proposed method can successfully handle the location uncertainties and pose variances. We provide more visualization results in Figure \ref{fig:supp-mask}.

To some extent, the methods relying on attention masks and rigid parts are at two extremes.
The former attempts to completely cover the informative pixels in a highly adaptive way, but mostly fails since we have only image-level annotations.
The latter one just totally discards the adaptive factors, which are less robust to pose variances.
Therefore, the proposed method attempts to achieve a balance between these two extremes, by constraining the attentional regions to several bounding boxes, which relatively coarse but more interpretable and controllable.

\begin{table}[t]
  \small
  \begin{center}
    \begin{tabular}{|l|cc|}
      \hline
      \backslashbox{Method}{Metric} &mA &F1 \\ \hline
      Rigid Part &76.56 &78.84 \\
      Attention Mask &78.35 &79.51 \\
      \hline \hline
      \textbf{Attribute Region} &\multicolumn{1}{l}{\textbf{81.87}} &\multicolumn{1}{l|}{\textbf{80.16}}\\ \hline
    \end{tabular}
  \end{center}
  \caption{Experimental results of different attribute-specific localization methods evaluated on RAP dataset.}
  \label{Tab:mask}
\end{table}

\begin{table*}[t]
  \small
  \begin{center}
  \begin{tabular}{l|ccccc||ccccc|cc}
    \hline
    Dataset & \multicolumn{5}{c||}{PETA} & \multicolumn{7}{c}{RAP} \\ \hline
    \backslashbox{Method}{Metric} &mA &Accu &Prec &Recall &F1 &mA &Accu &Prec &Recall &F1 &\#P &GFLOPs \\ \hline
    ACN \cite{sudowe2015person} &81.15 &73.66 &84.06 &81.26 &82.64 &69.66 &62.61 &80.12 &72.26 &75.98 &- &- \\
    DeepMar \cite{deepmar} &82.89 &75.07 &83.68 &83.14 &83.41 &73.79 &62.02 &74.92 &76.21 &75.56 &58.5M &\textbf{0.72} \\ \hline
    JRL \cite{wang2017attribute} &85.67 &- &86.03 &85.34 &85.42 &77.81 &- &78.11 &78.98 &78.58 &- &- \\
    \text{JRL*} \cite{wang2017attribute} &82.13 &- &82.55 &82.12 &82.02 &74.74 &- &75.08 &74.96 &74.62 &- &- \\
    GRL \cite{zhao2018grouping} &\textbf{86.70} &- &84.34 &\textbf{88.82} &86.51 &81.20 &- &77.70 &80.90 &79.29 &$>$50M &$>$10 \\ \hline
    HP-Net \cite{hpnet} &81.77 &76.13 &84.92 &83.24 &84.07 &76.12 &65.39 &77.33 &78.79 & 78.05 &- &- \\
    VeSPA \cite{deepview} &83.45 &77.73 &86.18 &84.81 &85.49 &77.70 &67.35 &79.51 &79.67 &79.59 &17.0M &$>3$ \\
    DIAA \cite{deepimb} &84.59 &78.56 &86.79 &86.12 &86.46 &- &- &- &- &- &- &- \\ \hline
    PGDM \cite{li2018pose} &82.97 &78.08 &\textbf{86.86} &84.68 &85.76 &74.31 &64.57 &78.86 &75.90 &77.35 &87.2M &$\approx$1 \\
    LG-Net \cite{liu2018localization} &- &- &- &- &- &78.68 &68.00 &\textbf{80.36} &79.82 &80.09 &$>$20M &$>4$ \\ \hline\hline
    BN-Inception &82.66 &77.73 &86.68 &84.20 &85.57 &75.76 &65.57 &78.92 &77.49 &78.20 &\textbf{10.3M} &1.78 \\
    \textbf{Ours} &\multicolumn{1}{c}{86.30} &\multicolumn{1}{c}{\textbf{79.52}} &\multicolumn{1}{c}{85.65} &\multicolumn{1}{c}{88.09} &\multicolumn{1}{c||}{\textbf{86.85}} &\multicolumn{1}{c}{\textbf{81.87}} &\multicolumn{1}{c}{\textbf{68.17}} &\multicolumn{1}{c}{74.71} &\multicolumn{1}{c}{\textbf{86.48}} &\multicolumn{1}{c}{\textbf{80.16}} &\multicolumn{1}{|c}{17.1M} &\multicolumn{1}{c}{1.95} \\ \hline
  \end{tabular}
  \end{center}
  \caption{Quantitative comparisons against previous methods on PETA and RAP datasets.
  We divide these methods into four groups: holistic methods, relation-based methods, attention-based methods, and part-based methods, from top to bottom.
  \text{JRL*} is the single model version of JRL.
  The precision and recall metrics are not so reliable in class-imbalanced datasets while the mA and F1 score are more convictive.
  Best results are in \textbf{bold}.
  For RAP dataset, we further provide comparisons on the number of parameters (\#P) and complexity (GFLOPs).
  }
  \label{Tab:peta-rap}
\end{table*}

\begin{table}[t]
  \begin{center}
  \small
  \begin{tabular}{l|ccccc}
    \hline
    Dataset &\multicolumn{5}{c}{PA-100K} \\ \hline
    Method &mA &Accu &Prec &Recall &F1 \\ \hline
    DeepMar \cite{deepmar} &72.70 &70.39 &82.24 &80.42 &81.32 \\
    HP-Net \cite{hpnet} &74.21 &72.19 &82.97 &82.09 &82.53 \\
    PGDM \cite{li2018pose} &74.95 &73.08 &84.36 &82.24 &83.29 \\
    VeSPA \cite{deepview} &76.32 &73.00 &84.99 &81.49 &83.20 \\
    LG-Net \cite{liu2018localization} &76.96 &75.55 &\textbf{86.99} &83.17 &85.04 \\ \hline\hline
    BN-Inception &77.47 &75.05 &86.61 &85.34 &85.97 \\
    \textbf{Ours} &\multicolumn{1}{c}{\textbf{80.68}} &\multicolumn{1}{c}{\textbf{77.08}} &\multicolumn{1}{c}{84.21} &\multicolumn{1}{c}{\textbf{88.84}} &\multicolumn{1}{c}{\textbf{86.46}} \\ \hline
  \end{tabular}
  \end{center}
  \caption{Quantitative comparisons on PA-100K dataset.}
  \vspace{-3mm}
  \label{Tab:pa100k}
\end{table}

\subsection{Comparison with State-of-the-art Methods}
In this subsection, we compare the performance of our proposed method against several state-of-the-art methods.
As mentioned in Section \ref{sec:related-works}, we divide these methods into four categories:
(1) Holistic methods including ACN \cite{sudowe2015person} and DeepMar \cite{deepmar}, which first take CNN to jointly learn multiple attributes.
(2) Relation-based methods including JRL \cite{wang2017attribute} and GRL \cite{zhao2018grouping}, which both exploit the semantic relations by a CNN-RNN based model.
(3) Attention-based methods including HP-Net \cite{hpnet} and DIAA \cite{liu2018localization} relying on multi-scale attention mechanism, and VeSPA \cite{deepview} which perform view-specific attribute prediction through a coarse view predictor.
(4) Part-based methods including recently proposed PGDM \cite{li2018pose} and LG-Net \cite{liu2018localization}, which relying on external pose estimation or region proposal module.

Table \ref{Tab:peta-rap} and Table \ref{Tab:pa100k} show the comparison results on three different datasets.
The results suggest that our proposed method achieves superior performances compared with existing works under both label-based and instance-based metrics on all three datasets.
Compared with the previous methods relying on attribute-agnostic attention or extra part localization mechanism, the proposed method can achieve a significant improvement across all datasets, which demonstrates the effectiveness of attribute-specific localization.
Although a slightly lower mA score is achieved than the relation-based method GRL on PETA dataset, due to their stronger Inception-v3 backbone network (with twice as many parameters as ours),
%
we can still outperform them on other metrics and datasets. 
On the more challenging dataset PA-100K, the proposed method largely outperforms all previous works, improving $3.7\%$ and $1.4\%$ in mA and F1, respectively, over the second best results.
Notably, the proposed method surpasses the baseline model with a significant margin, especially on the label-based metric mA ($3.6\%$, $6.1\%$, and $3.2\%$ on three datasets, respectively).
Note that the proposed method often achieve a lower precision but higher recall, while these two metrics are not so reliable, especially in class-imbalanced datasets.
Moreover, the two metrics are inversely correlated, \ie, increase in one metric always leads to decrease in another (\eg, by modulating the class weights in the loss function).
The mA and F1 metrics are more appropriate in measuring the performance of an attribute recognition model.
Our method consistently achieves the best results in these two metrics.

We provide a comparison of the computational cost for different methods (rightmost columns in Table \ref{Tab:peta-rap}) on RAP dataset.
For the number of parameters, theoretically, there are totally $(\frac {C ^ { 2 } } { 8 } + 4C)$ trainable parameters in each ALM:
$4C$ from the STN module, $\frac {C ^ { 2 } } { 8 }$ from the channel-attention module, where $C$ is the number of input channels.
As shown, the proposed model has much fewer trainable parameters than previous models.
In terms of model complexity, even with 51 attributes, the proposed model is still light-weight as only 0.17 GFLOPs are added to the backbone network.
The reason is that ALM contains only FC-layers (or 1$\times$1 Conv), which involves much fewer FLOPs than 3$\times$3 Conv-layers.
In general, the entire model is much more efficient than previous models.

\section{Conclusion}
We propose an end-to-end framework for pedestrian attribute recognition, which can automatically localize the attribute-specific regions at multiple feature levels.
Moreover, we apply a feature pyramid architecture to enhance the attribute localization and region-based feature learning in a mutually reinforcing manner.
Experimental results on PETA, RAP, and PA-100K datasets show that the proposed method can significantly outperform most of the existing methods.
The extensive analysis suggests that the proposed method can successfully localize the most informative region for each attribute in a weakly-supervised manner.

\noindent\textbf{Acknowledgements} \
This work was supported in part by the National Key Research and Development Program of China under Grant 2017YFA0700904, in part by the National Natural Science Foundation of China under Grant 61836014, and Grant 61620106010.

{\small
\bibliographystyle{ieee_fullname}
\bibliography{egbib}
}

\appendix

\section*{Appendix}

\setcounter{table}{0}
\setcounter{figure}{0}
\setcounter{equation}{0}
\renewcommand{\theequation}{S\arabic{equation}}
\renewcommand{\thefigure}{S\arabic{figure}}
\renewcommand{\thetable}{S\arabic{table}}
\renewcommand{\theHfigure}{S\arabic{figure}}
\renewcommand{\theHtable}{S\arabic{table}}

\section{Implementation Details} \label{appendix_a}
We adopt the BN-Inception model pretrained from ImageNet as the backbone network.
The proposed framework is implemented with PyTorch framework and trained end-to-end with only image-level annotations.
We adopt Adam optimizer since it converges faster than SGD in our experiments with momentum set to $0.9$ and a weight decay equals to $0.0005$.
The initial learning rate equals to $0.0001$ and the batch size is set to $32$.
For RAP and PA-100K dataset, we train the model for 30 epochs and the learning rate decays by $0.1$ every $10$ epochs.
For the smaller PETA dataset, we double the training epochs.
For data preprocessing, we resize the input pedestrian images to $256 \times 128$ and apply random horizontal mirroring and data shuffling for data augmentation.

\section{Different Attribute-Specific Methods} \label{appendix_b}
In Section \ref{sec4_4}, we compare the proposed method against the other two attribute-specific localization methods, including visual attention and rigid parts.
Different from most existing attribute-agnostic attention-based and part-based methods, we build two attribute-specific models based on these ideas for comparison.
Here we show the details of the compared models.

\begin{figure}[h]
\begin{center}
  \includegraphics[width=1.0\linewidth]{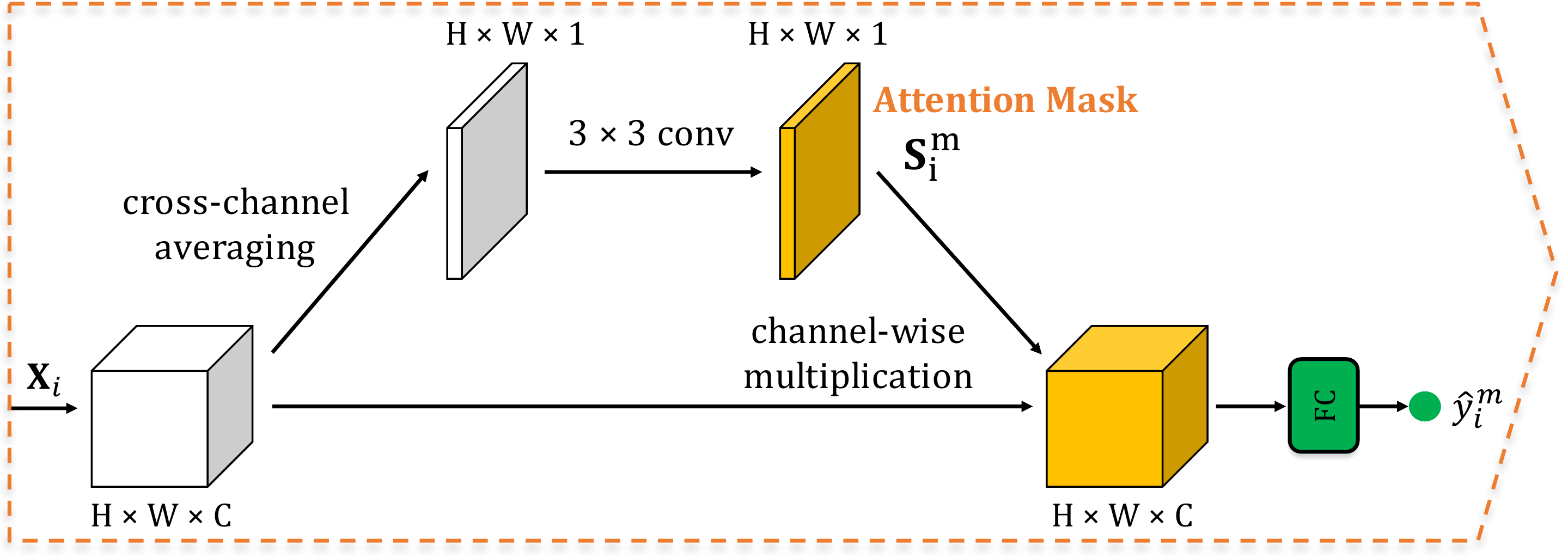}
\end{center}
   \caption{Details of the spatial attention module for one attribute at a singe level.
   The expected attention mask follows a cross-channel averaging layer and a $3 \times 3$ Conv-BN-ReLU block.
   }
\label{fig:supp-att}
\end{figure}

\textbf{Attention Masks Model}. We replace the proposed ALM with a spatial attention module while keeping others unchanged for fair comparison.
The spatial attention module is implemented by a tiny 3-layers sub-network, as shown in Figure \ref{fig:supp-att}, which is inspired by HA-CNN \cite{li2018harmonious}.
The input features $\mathbf{X}_{i} \in \mathbb{R}^{H \times W \times C}$ at the $i$-th level (a certain layer in the backbone network, totally three levels) are first fed into a cross-channel averaging layer.
A $3\times3$ Conv-BatchNorm-ReLU block is followed to generate the expected attention mask $\mathbf{S}^{m}_{i} \in \mathbb{R}^{H \times W \times 1}$, which is used for localizing the $m$-th attribute at the $i$-th level.
All channels share the identical spatial attention mask.
Subsequently, the attentive features are obtained by channel-wise multiplying the attention mask with the input features,
and the corresponding prediction is calculated as follows:
\begin{equation}
  \hat y^{m}_{i} = f(\mathbf{S}^{m}_{i} \cdot \mathbf{X}_{i}),
\end{equation}
where $f$ denotes a fully-connected layer.
Each spatial attention module only serves one attribute at a singe level, the same as Figure \ref{fig:ALM}.

\begin{figure}[h]
\begin{center}
  \includegraphics[width=0.9\linewidth]{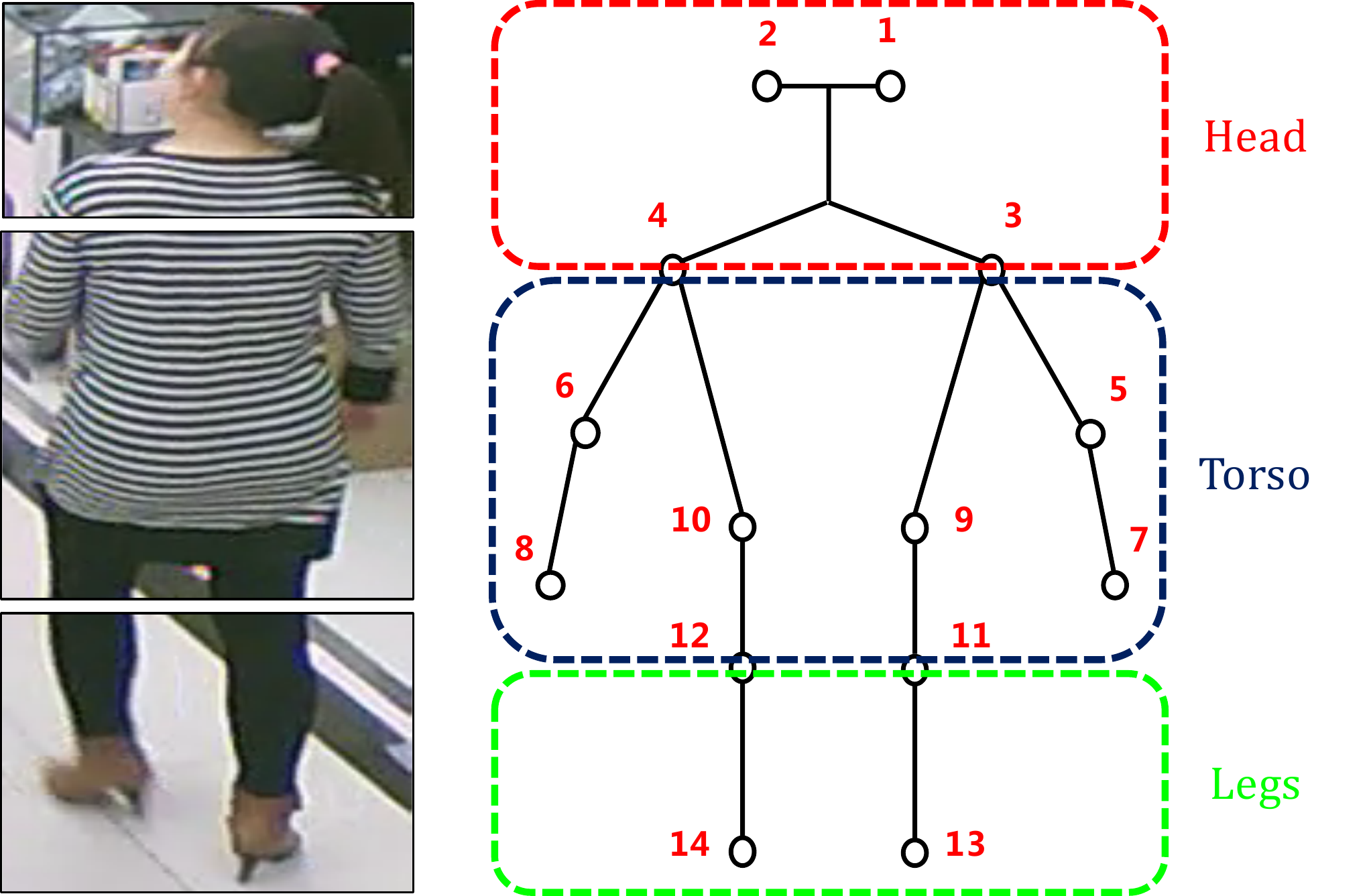}
\end{center}
   \caption{Illustration of body-parts generation. We divide a pedestrian image into three body-part regions (\textit{head, torso, and legs}) based on 14 human body keypoints.}
\label{fig:supp-roi}
\end{figure}

\begin{figure}[t]
\begin{center}
  \includegraphics[width=1.0\linewidth]{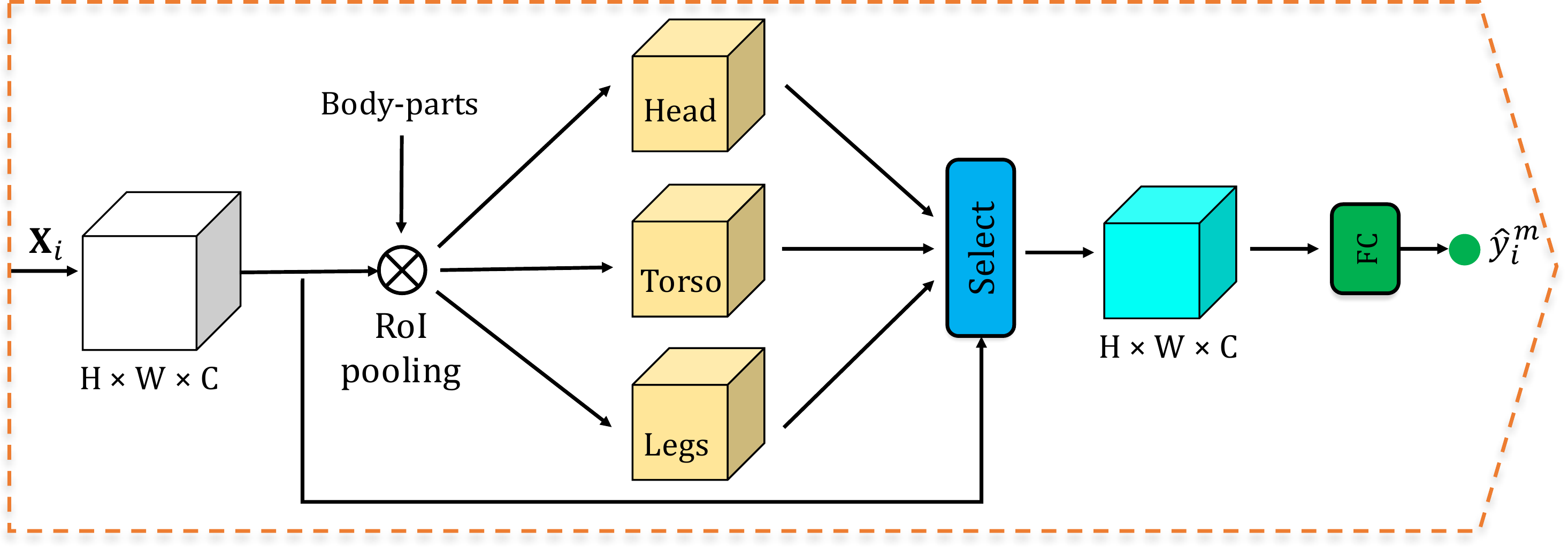}
\end{center}
   \caption{Details of the body-parts guided module  for one attribute at a singe level.
   The three body-part regions are calculated based on several human body keypoints predicted by a pretrained pose estimation model.
   The local features belonging to different body-parts are extracted by an RoI pooling layer.
   The most relevant features are selected for attribute classification according to the predefined attribute-region correspondence (Table \ref{Tab:attri-part}).
   }
\label{fig:supp-part}
\end{figure}

\textbf{Rigid Parts Model}. For attribute-specific part-based model, we replace ALM with a body-parts guided module, as shown in Figure \ref{fig:supp-part}.
The key idea is to associate each attribute with a predefined body region, including \textit{head, torso, legs}, and the whole image, \eg, the \textit{LongHair} attribute is associated with the head part.
Since the body-part annotations are unavailable on most pedestrian attribute datasets, we adopt an external pose estimation model to localize the body parts, which is inspired by SpindleNet \cite{Zhao_2017_CVPR}.
Specifically, we localize 14 human body keypoints for each pedestrian image using a pretrained pose estimation model \cite{Zhao_2017_CVPR}.
The pedestrian image is then divided into three body-part regions based on these keypoints, as shown in Figure \ref{fig:supp-roi}.
In the body-parts guided module (Figure \ref{fig:supp-part}), the body-part-based local features are extracted from the input features $\mathbf{X}_{i}$ through an RoI pooling layer \cite{girshick2015fast}.
For attribute prediction, the most relevant features are selected according to the attribute-region correspondence, as listed in Table \ref{Tab:attri-part}, \eg recognizing \textit{hat} using features only from the \textit{head} part.

\begin{table}[t]
  \footnotesize
\begin{center}
\begin{tabular}{|c|l|}
  \hline
Region &\multicolumn{1}{c|}{Attributes}                                                                                                                                                                  \\ \hline\hline
Head   & \begin{tabular}[c]{@{}l@{}}BaldHead, LongHair, BlackHair, Hat, Glasses, \\ Muffler, Calling\end{tabular}                                                                                       \\ \hline
Torso  & \begin{tabular}[c]{@{}l@{}}Shirt, Sweater, Vest, TShirt, Cotton, Jacket, \\ Suit-Up, Tight, ShortSleeve, LongTrousers, \\ Skirt, ShortSkirt, Dress, Jeans, TightTrousers, \\ CarryingbyArm, CarryingbyHand\end{tabular} \\ \hline
Legs   & \begin{tabular}[c]{@{}l@{}}LeatherShoes, SportShoes, Boots, ClothShoes, \\ CasualShoes\end{tabular}                                                                                   \\ \hline
Whole  & \begin{tabular}[c]{@{}l@{}}Female, AgeLess16, Age17-30, Age31-45, BodyFat,\\ BodyNormal, BodyThin, Customer, Clerk, Backpack,\\ SSBag, HandBag, Box, PlasticBag, PaperBag,\\ HandTrunk, OtherAttchment, Talking, Gathering,\\ Holding, Pushing, Pulling \end{tabular}    \\
  \hline
\end{tabular}
\end{center}
\caption{Attribute-region correspondence in RAP dataset.}
\label{Tab:attri-part}
\end{table}

\begin{figure*}[t]
\begin{center}
  \includegraphics[width=0.8\linewidth]{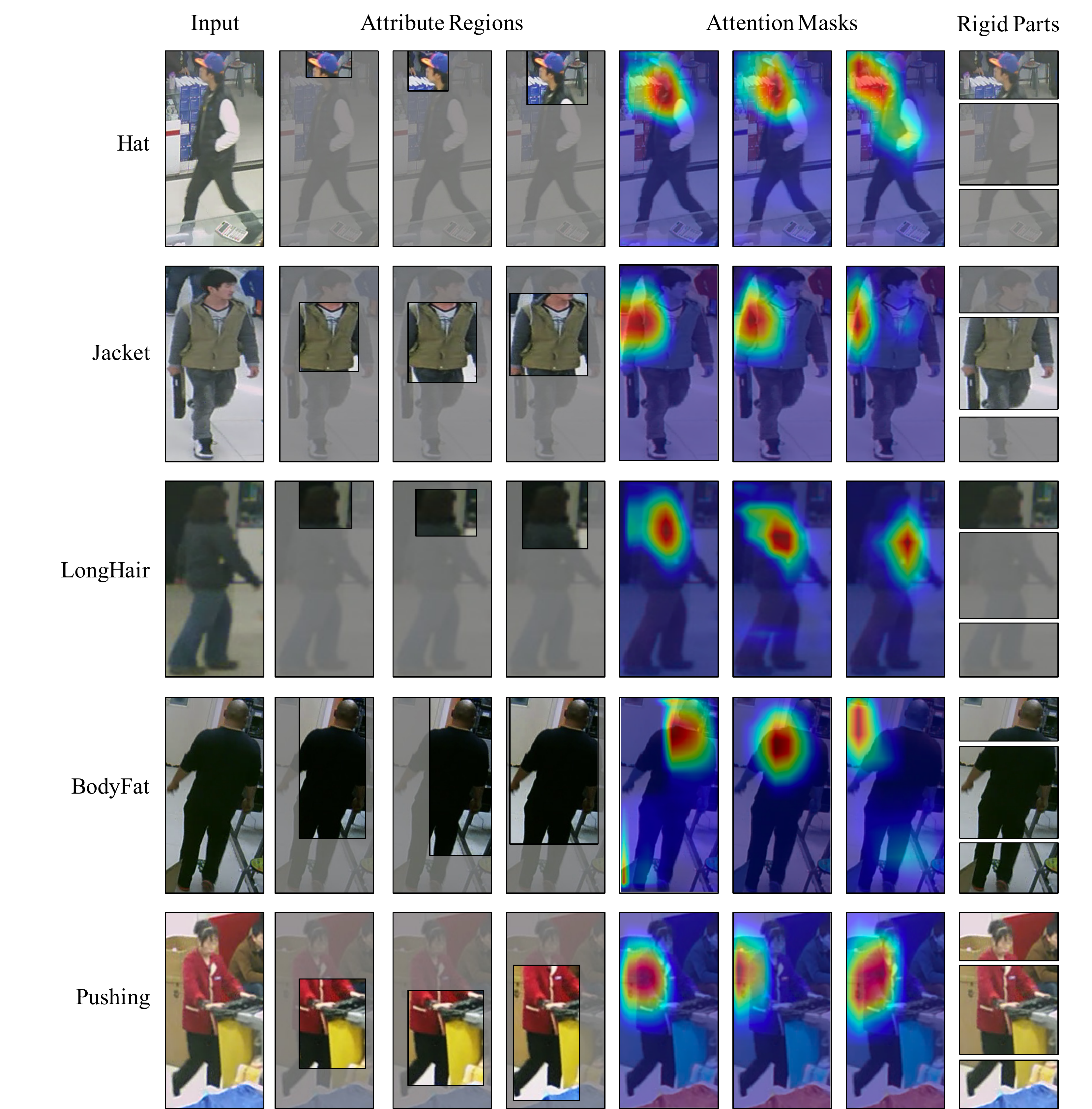}
\end{center}
   \caption{Case studies of different attribute-specific localization methods for five different attributes.}
\label{fig:supp-mask}
\end{figure*}

We provide more localization results belong to different attributes, as shown in Figure \ref{fig:supp-mask}.

\end{document}